# Improved Sensitivity of Base Layer on the Performance of Rigid Pavement

Sajib Saha, Ph.D.
Post-doctoral Research Associate
Texas A&M University System
3135 TAMU, CE/TTI Bldg. 503H, College Station, Texas 77843
Phone: (864) 986-2451, Email: sajibsaha@tamu.edu

Fan Gu, Ph.D., P.E., M.ASCE
Assistant Research Professor
National Center for Asphalt Technology
Auburn University
Phone: (334) 844-6251, Email: fzg0014@auburn.edu

Xue Luo, Ph.D., A.M. ASCE
College of Civil Engineering and Architecture
Zhejiang University
866 Yuhangtang Road, An-zhong Bldg.
Hangzhou 310058, China
Office: +86 571-88206542, Email: xueluo@zju.edu.cn

Robert L. Lytton, Ph.D., P.E. ,F.ASCE
Professor, Fred J. Benson Chair
Zachry Department of Civil Engineering
Texas A&M University
3136 TAMU, CE/TTI Bldg. 503A, College Station, Texas 77843
Phone: (979) 845-9964, Email: r-lytton@civil.tamu.edu

## ABSTRACT

The performance of rigid pavement is greatly affected by the properties of base/subbase as well as subgrade layer. However, the performance predicted by the AASHTOWare Pavement ME design shows low sensitivity to the properties of base and subgrade layers. To improve the sensitivity and better reflect the influence of unbound layers a new set of improved models i.e., resilient modulus ($M_R$) and modulus of subgrade reaction (*k*-value) are adopted in this study. An Artificial Neural Network (ANN) model is developed to predict the modified k-value based on finite element (FE) analysis. The training and validation datasets in the ANN model consist of 27000 simulation cases with different combinations of pavement layer thickness, layer modulus and slab-base interface bond ratio. To examine the sensitivity of modified $M_R$ and *k*-values on pavement response, eight pavement sections data are collected from the Long-Term Pavement performance (LTPP) database and modeled by using the FE software ISLAB2000. The computational results indicate that the modified $M_R$ values have higher sensitivity to water content in base layer on critical stress and deflection response of rigid pavements compared to the results using the Pavement ME design model. It is also observed that the *k*-values using ANN model has the capability of predicting critical pavement response at any partially bonded conditions whereas the Pavement ME design model can only calculate at two extreme bonding conditions (i.e., fully bonding and no bonding).

# 1. Introduction

Rigid pavement structure typically consists of a Portland cement concrete (PCC) surface layer, and an intermediate base course layer along with the underlying subgrade layer. It is recognized that the overall performance of rigid pavement is significantly affected by base and subgrade layers [1-4]. For design and maintenance of new and existing pavements, the American Association of State Highway and Transportation Officials (AASHTO) currently follows a design guide named AASHTOWare Pavement Mechanistic-Empirical (ME) Design which provides an ME methodology for the analysis and performance prediction of new pavements and overlays. However, recent studies have indicated that the performance predicted by the design guide shows low sensitivity to the properties of base/subgrade layers [5]. The properties of base and subgrade layers included in the Pavement ME Design guide are: resilient modulus ($M_R$), soil-water characteristics curve (SWCC), thickness, erodibility index, load transfer efficiency (LTE), slab-base interface bond, and ground water depth [6]. To improve the sensitivity and thereby achieve the realistic influence of the base/subgrade layer properties on the performance of rigid pavement, it is required to enhance the property models and implement them in the Pavement ME Design guide. A new set of $M_R$ and modulus of subgrade reaction ($k$-value) models have been proposed by Lytton and Saha et al. respectively to improve moisture and slab-base interface bond sensitivity in corresponding models [7-8]. These models are adopted in this study for evaluation of the material sensitivity on rigid pavement performance.

As a fundamental material property of base and subgrade layers $M_R$ represents the load carrying capacity under different conditions i.e., water content, density, and stress level [9]. The most generalized $M_R$ model is developed in the National Cooperative Highway Research Program (NCHRP) Project 1-28A [10].

$$M_R = k_1 P_a (\frac{I_1}{P_a})^{k_2} (\frac{\tau_{oct}}{P_a}+1)^{k_3} \tag{1}$$

where $I_1$ is the bulk stress ($\sigma_1+\sigma_2+\sigma_3$); $\tau_{oct}$ is the octahedral shear stress; $P_a$ is the atmospheric pressure; and $k_1$, $k_2$ and $k_3$ are the material parameters. To incorporate the moisture dependency in resilient modulus, the AASHTO developed an empirical $M_R$ model that is later adopted by the AASHTOWare Pavement ME Design [11].

$$\log \frac{M_R}{M_{Ropt}} = a + \frac{b-a}{1+\exp[\ln\frac{-b}{a}+k_m(S-S_{opt})]} \tag{2}$$

where $M_{Ropt}$ is the resilient modulus at a reference condition; $a$, $b$, $k_m$ are the fitting parameters; S is the degree of saturation; and $S_{opt}$ is the saturation at optimum water content.

However, other researchers have shown that the pavement performance indicators have extremely low sensitivity to the degree of saturation indicated by ($S$-$S_{opt}$) in Equation 2 [12]. The reasons for such problems are illustrated in [13], and they suggested that for the same value of ($S$-$S_{opt}$), the variation of the matric suction can be substantial from one material to another. Therefore, the resilient modulus can change significantly for the different materials with the identical value of ($S$-$S_{opt}$). As a result, the inclusion of the degree of saturation alone cannot accurately reflect the change of resilient modulus due to water content. Thereby, both the degree of saturation and the matric suction should be incorporated in the resilient modulus model for unbound layer and subgrade. There have been a number of studies recognizing that matric suction plays a significant role in the $M_R$ characteristics [14-18]. Therefore, to consider the influence of both water content and suction in $M_R$ characteristics, Lytton proposed a modified $M_R$ model [7].

$$E_y = k_1 P_a \left(\frac{I_1 - 3\theta f h_m}{P_a}\right)^{k_2} \left(\frac{\tau_{oct}}{P_a}\right)^{k_3} \tag{3}$$

where $\theta$ is the volumetric water content; $h_m$ is the matric suction in the base matrix; $f$ is the saturation factor, $1 < f < 1/\theta$; and $k_1$, $k_2$ and $k_3$ are regression coefficients which depend upon material properties

The water content in the base and subgrade layers depends on the in-situ matric suction in the field. Water content and the corresponding suction reaches in an equilibrium condition several years after construction. In the Pavement ME Design Guide, the water content/degree of saturation in the underlying layers is calculated internally using the embedded Enhanced Integrated Climatic Model (EICM). The EICM model relies on a correlation between the degree of saturation and the depth of water table to calculate the in-situ water content of base and subgrade layers. However, recent study has demonstrated that if there is no water table data available or water table is below 10 m, the water level will be determined by the equilibrium suction and its variations [19]. Therefore, the equilibrium suction map developed by Saha et al. is adopted in this study for determination of realistic water content range and suction value in base/subgrade layers [19] and then used in the $M_R$ model.

Similarly, the $k$-value is another important property to evaluate the subgrade strength and design rigid pavements [8]. The $k$-value in the Pavement ME Design Guide is called effective dynamic $k$-value that involves the subgrade $M_R$ value and deflection pattern of pavement surface but neglects the shear interaction within the supporting media. The slab-base interface is assumed to be only at two extreme bonded conditions, i.e., full bond and no bond. However, none of these assumptions are found to be realistic in the field [20]. A more realistic assumption of interfacial bonding can infer a more accurate prediction of pavement performance [21-23]. To solve this problem, Saha et al. developed a modified $k$-value model that considers the various degrees of slab-base interface bond [8], which is used in this study.

The current AASHTOWare Pavement ME Design utilizes the built-in performance models for rigid pavements based on the slab and base properties and subgrade effective dynamic $k$-value. Therefore, it is required to develop a methodology to incorporate the modified $k$-value in Pavement ME Design and further improve the sensitivity of interface bonding on predicted performance. In this study, an Artificial Neural Network (ANN) model is used to predict the modified $k$-value for a wide range of pavement layer thickness, modulus and interface bonding ratios. The incorporation of ANN model to Pavement ME Design is much easier than replacing the modified analytical model. Moreover, the calculation of modified $k$-value requires numerical modeling of pavement structure to simulate the slab-base interface bond and further estimate the deflection basin. These numerical computations are time consuming and need commercial finite element (FE) software. Thus, the use of ANN approach will also eliminate the complexity of FE analysis for pavement structure.

The primary objective of this paper is to develop an ANN model to take account of the influences of water content and interface bond on rigid pavement performance. The following section describes the construction of the ANN model to predict modified $k$-value from a wide range of pavement layer thickness, modulus and interface bonding ratios. After that, eight pavement sections are selected from the Long-Term Pavement Performance (LTPP) database and are used to validate the ANN model predictions with FE simulation results. The sensitivity of water content and degree of bonding on *k*-value from ANN model are evaluated and compared against the predicted *k*-values using the Pavement ME design model. Subsequently, the sensitivity of water content and degree of bonding are evaluated on critical rigid pavement responses and thereby project the overall performance. The concluding section summarizes the significant findings of this study.

## 2. Development of ANN model to predict modified *k*-value

The advantage of the ANN approach is its ability to process the data information of complex systems and adapt the system parameters accordingly. An ANN model consists of many inter-connected neurons which establish the correlations between the input variables $X_i$ and the output variables $Y_j$. The input variables $X_i$ and the output variables $Y_j$ are usually normalized to $x_i$ and $y_j$, respectively. The normalized input parameters $x_i$ are connected to each neuron through the weight factor, $w_{ji}$, as shown in Equation 4.

$$y_j = f\left(\sum_{i=1}^{n} w_{ji} x_i\right) + b_j \tag{4}$$

where $f$ is a transfer function, which can take any value between plus and minus infinity as the input, and converts the output in the range of 0 to 1. The weight factors $w_{ji}$ and bias $b_j$ in Equation 4 are adjusted based on the minimum error function.

The ANN approach is increasingly used in pavement engineering to develop prediction models of complex input-output dependencies. These prediction models are developed based on a large number of data collected from experiments or numerical analyses [9, 18, 24]. In this study, the FE program ABAQUS is used to calculate the modified *k*-value for a wide range of pavement layer combinations and construct the training dataset. The development of ANN model includes two critical steps: 1) calculation of modified *k*-value based on the FE analysis; and 2) construction of the ANN architecture.

## 2.1. Calculation of modified k-value based on FE analysis

The determination of the modified *k*-value of a specific pavement structure requires to estimate the pavement structural response (i.e., surface deflection basin) to the falling weight deflectometer (FWD) loading. Herein, the surface deflection basin is computed using the FE program ABAQUS. Pavement structures are modeled with different combinations of pavement layer thicknesses, layer modulus and slab-base interface bond. Figure 1 shows a typical rigid pavement structure and the corresponding finite element model.

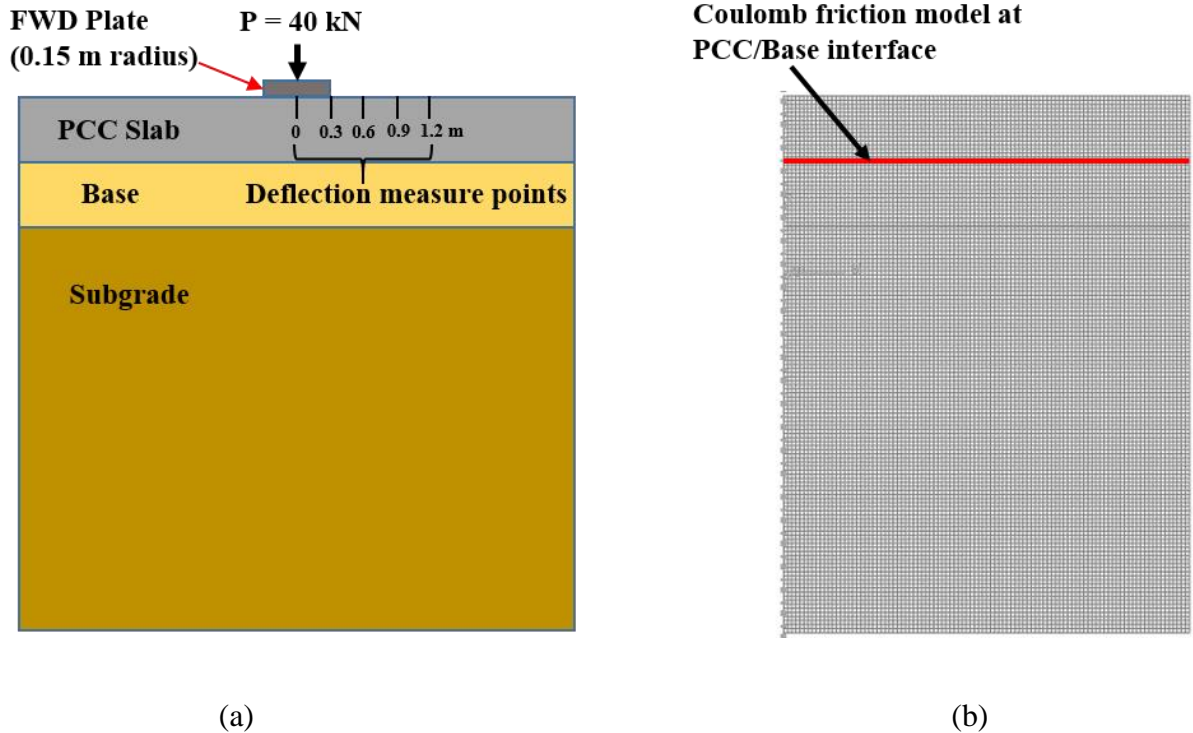

(a) (b)

**Figure 1. (a) Illustration of a typical rigid pavement structure; (b) finite element model of pavement in ABAQUS**

In order to properly reflect the effect of interface bond strength on the overall pavement structural response, it is necessary to accurately characterize the slab-base interface. The contact interaction model in ABAQUS has a normal and a tangential contact stress mechanism across the interface between contacting bodies. The Coulomb friction model is used in this study to characterize the tangential behavior of slab-base interface. The Coulomb friction model defines the maximum shear stress, $(\tau_{zx})_{2\theta f}$, at which sliding of the surfaces starts on the slab base interface. The input parameter, $\mu$, defined as the coefficient of friction, is expressed as shown in Equation 5.

$$\mu = \frac{N}{(\tau_{zx})_{2\theta_f}} \quad (5)$$

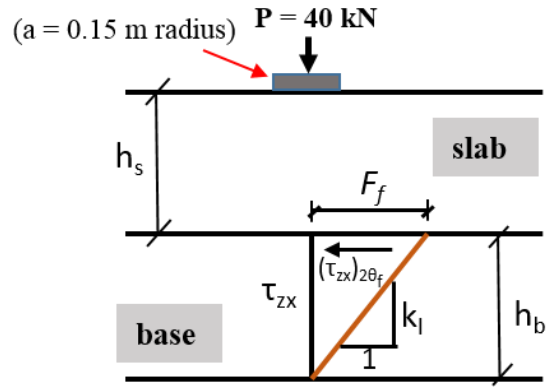

**Figure 2. Illustration of the maximum shear stress and maximum elastic slip displacement at slab-base interface**

Note that, the vertical applied pressure, $N$ is determined using the Timoshenko analytical solutions for the stress state in a half space [25].

$$N = \frac{3Ph_s^3}{2\pi(h_s^2 + a^2)^{5/2}} \quad (6)$$

where $P$ is the applied load, $h_s$ is the thickness of slab and $a$ is the radial distance from loading point.

The term $(\tau_{zx})_{2\theta_f}$ in Equation 7 is expressed as follows [8].

$$(\tau_{zx})_{2\theta_f} = \delta * \frac{3P}{2\pi} \left[ \frac{ah_s^2}{(h_s^2+a^2)^{5/2}} \right] \frac{h_b \left( h_s + \frac{h_b}{2} - \bar{z} \right)}{h_s(\bar{z} - \frac{h_s}{2})} \quad (7)$$

where $h_b$ is the thickness of the base; $\bar{z}$ is the distance of neutral axis from point of interest; and δ is the degree of bonding on slab-base interface.

The contact model applies a stiffness method that allows a small motion of the surfaces called "elastic slip" when they adhere together. The magnitude of the elastic slip displacement, $F_f$, as shown in Figure 2, is calculated as follows

$$F_f = \frac{(\tau_{zx})_{2\theta_f}}{k_l} \quad (8)$$

where $k_l$ is the horizontal shear stiffness, expressed as follows

$$k_l = \frac{\pi G}{l\upsilon} \quad (9)$$

where $G = \frac{E}{2(1+\upsilon)}$ is the shear modulus; and $l = \left( \frac{4h_b^3 \upsilon}{\pi} \right)^{1/3}$ is the radius of relative shear stiffness.

The slab-base interaction model in ABAQUS can accurately take account for the user defined degree of bonding which quantifies the coefficient of friction, $\mu$, and the elastic slip displacement, $F_f$. Therefore, the change in interface bonding ratio causes the change in calculated deflection basins and the corresponding modified k-values. The FWD sensor deflections (0 cm,

30.48 cm, 60.96 cm, and 91.44 cm away from the loading point) are obtained from the FE analysis, and the basin area, BA, is calculated as:

$$BA = \frac{SS}{2 \times D_0}[D_0 + 2(D_1 + ....... + D_{j-1}) + D_j] \tag{10}$$

where $SS$ is the FWD sensor spacing (30.48 cm ≈ 12 inch); and $D_j$ is the surface deflection at the location of sensors ($j$=0 to 3).

The effective relative stiffness length is calculated based on the basin area, BA, as follows:

$$l_e = [\frac{\ln(\frac{k_1 - BA}{k_2})}{-k_3}]^{1/k_4} \tag{11}$$

where $k_1$, $k_2$, $k_3$ and $k_4$ are the field correlation coefficients, $k_1$=36, $k_2$=1812.597, $k_3$=2.559, and $1/k_4$=4.387 [26]

The effective relative stiffness length is related to dimensionless deflection coefficient ($d^*$). The dimensionless deflection coefficient for any sensor is determined as follows,

$$d_r^* = a.e^{-b.e^{-c.l_e}} \tag{12}$$

where $a$, $b$ and $c$ are the regression coefficients, $a = 0.12450$, $b = 0.14707$ and $c = 0.07565$ at loading point [26].

Finally, the modified $k$-value is calculated based on the sensor deflection using Equation 13

$$k = \frac{P.d_r^*}{d_r.l_e^2} \tag{13}$$

where $P$ is the applied load (40 kN ≈ 9000 lbf); $d_r$ is the calculated deflection at the loading point using analysis.

## 2.2. Construction of ANN architecture

A three-layered ANN architecture consisting of one input layer, one hidden layer and one output layer is constructed in this study to predict the modified $k$-value from design inputs. As shown in Figure 3, pavement layer thicknesses (i.e., slab and base thicknesses), layer modulus (i.e.; slab, base and subgrade moduli), and slab-base interface bonding ratio are introduced as input parameters. These parameters in the ANN model are selected based on the sensitivity analysis conducted by Saha et al. [27].

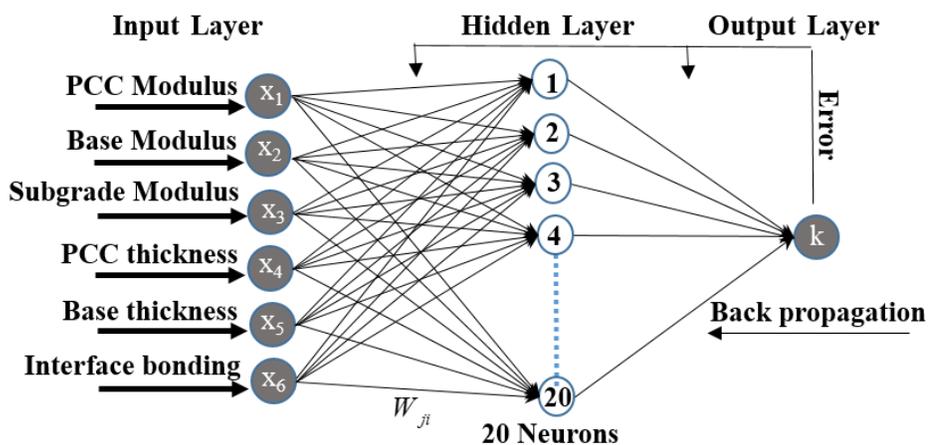

**Figure 3. Architecture of three-layered neural network model for *k*-values**

Table 1 lists all the input parameters and their values in the ANN model. A total of 27,000 simulation cases are developed with the different combinations of input values. Several network topologies (6-10-1, 6-15-1, 6-20-1, 6-25-1, 6-10-10-1) with different number of layers and nodes are investigated. In this study, the performance of the hidden layer structures are determined based on their converged root mean square error (RMSE) values and the required computational times for training. The training results show that the structure 6-20-1 is converged to a desired RMSE value of 37.58 with a computational time of 62 sec. Note that the structures consisted of two hidden layers generate greater RMSE values and require more computational time than single hidden layered structure. Tables 2 lists all the trial neural network structures that are used to predict modified $k$-value and their corresponding RMSE and computational time.

**Table 1. Selected range of input parameters in ANN training dataset**

| Input parameters | Level | Input values |
|---|---|---|
| PCC thickness (mm) | 5 | 178, 216, 254, 292 and 330 |
| Base thickness (mm) | 5 | 89, 127, 165.1, 203.2 and 254 |
| PCC modulus (MPa) | 6 | 20784, 31026, 41368, 51710, 62052 and 75842 |
| Base modulus (MPa) | 6 | 34.5, 690, 1724, 3447, 5171 and 6895 |
| Subgrade modulus (MPa) | 6 | 34.5, 69, 138, 276, 414 and 551 |
| PCC-base interface bonding | 5 | 0, 0.25, 0.5, 0.75 and 1 |

**Table 2. List of ANN structures and the converged RMSE values and required computational time**

| ANN structure | RMSE | Computational time (sec) |
|---|---|---|
| 6-10-1 | 57.37 | 30 |
| 6-15-1 | 45.22 | 47 |
| 6-20-1 | 37.58 | 62 |
| 6-25-1 | 35.62 | 92 |
| 6-5-5-1 | 77.05 | 51 |
| 6-10-10-1 | 68.26 | 176 |

Note: Computer Info - 2.70 GHz Core i7 CPU with 16 GB RAM

To accurately determine the weight factors $w_{ji}$ of hidden neurons, the ANN model randomly divide the total dataset into two major parts: training and validating. The training data set is used to determine the trial weight factors, $w_{ji}$ and bias term, $b_j$, and the validating data set is employed to check the statistical accuracy and avoid the overfitting of the model predictions. In this study, the ratio of training and validation dataset is set to 4 to 1. The training function involves a global optimization algorithm to optimize the network performance function i.e., mean square error (MSE). The optimization algorithm calculates the gradient of the MSE with respect to the weight factors, $w_{ji}$, after each trial and update the weight and bias terms backward through the network. Matlab simulations are performed to determine the best training function for this dataset. Table 3 lists all the training functions that are tested [28] and their performance on modified $k$-value prediction. The simulation results show that the Levenberg-Marquardt Backpropagation function yields the lowest RMSE value, which is thereby selected for the ANN model.

**Table 3. Comparison of model performance on *k*-value prediction with different training functions**

| Training function | Algorithm description | RMSE | $R^2$ |
|---|---|---|---|
| trainbfg | BFGS Quasi-Newton Backpropagation | Not converge | Not converge |
| trainrp | Resilient Backpropagation | 58.64 | 0.94 |
| trainlm | Levenberg-Marquardt Backpropagation | 37.58 | 0.97 |
| trainscg | Scaled Conjugate Gradient Backpropagation | 64.57 | 0.93 |
| traincgb | Conjugate Gradient Backpropagation with Powell/Beale restarts | 57.45 | 0.94 |
| traincgf | Fletcher-Powell Conjugate Gradient Backpropagation | 64.28 | 0.93 |
| traincgp | Polak-Ribiére Conjugate Gradient Backpropagation | 65.3 | 0.92 |
| trainoss | One Step Secant Backpropagation | 69 | 0.91 |
| traingdx | Variable Learning Rate Backpropagation | Not converge | Not converge |
| traingdm | Gradient Descent with Momentum Backpropagation | 64.11 | 0.93 |
| traingd | Gradient Descent Backpropagation | 68.23 | 0.91 |

After the selection of the training function, it requires to select the proper learning function to embed in the training algorithm. While training function is a global algorithm that dictates all the weights and biases of a given network, the learning function are applied to adjust the weights and biases of individual neurons. Table 4 lists the available learning functions in Matlab and their RMSE values and computational time for the selected dataset. The results indicate that the learning function does not have any significant effect on the error precision, it only affects the computation efficiency of the ANN model. The learning function 'learngdm' is selected in this study because of the lowest computational time.

**Table 4. List of learning functions and comparison of their performance on *k*-value prediction**

| Learning function | Algorithm description | RMSE | Time (sec) |
|---|---|---|---|
| learncon | Bias Learning Rule | 37.55 | 78 |
| learngdm | Gradient Descent with Momentum Learning Rule | 37.58 | 62 |
| learnk | Kohonen Learning Rule | 37.45 | 129 |
| learnlv1 | LVQ1 Learning Rule | 37.51 | 70 |
| learnlv2 | LVQ2.1 Learning Rule | 37.53 | 67 |
| learnp | Perception Learning Rule | 37.55 | 85 |
| learnsomb | Batch Self-Organizing Map Learning Rule | 37.49 | 99 |
| Learnwh | Widrow-Hoff Learning Rule | 37.55 | 108 |

After determining the number of neurons in the hidden layer and training the ANN weights and biases, the network output (i.e., modified *k*-value) is expressed as a function of the input parameters $X = [x_1, x_2, x_3, x_4, x_5, x_6]^T$ and the network parameters. The network parameters are given by a combination of network weight matrix, as shown in Equation 14

$$W = \begin{pmatrix} w_{1,1} & w_{1,2} & w_{1,3} & w_{1,4} & w_{1,5} & w_{1,6} \\ w_{2,1} & w_{2,2} & w_{2,3} & w_{2,4} & w_{2,5} & w_{2,6} \\ \vdots & \vdots & \vdots & \vdots & \vdots & \vdots \\ w_{j,1} & w_{j,2} & w_{j,3} & w_{j,4} & w_{j,5} & w_{j,6} \end{pmatrix} \tag{14}$$

and the bias vector, $B = [b_1, b_2, b_3,\ldots\ldots\ldots\ldots, b_j]$ (15)

Herein, the scaler $j$ denotes the number of nodes in the hidden layer.

Finally, the predicted modified $k$-value can be calculated as a parametric nonlinear regression equation, as shown in Equations 16-18.

$$k = f_2\left(\sum_{j=1}^{j} f_1(w_{j,1}*x_1 + w_{j,2}*x_2 + w_{j,3}*x_3 + w_{j,4}*x_4 + w_{j,5}*x_5 + w_{j,6}*x_6 + b_j) + B_2\right) \tag{16}$$

$$f_1(x) = \frac{1}{1+\exp(-\varphi x)} \tag{17}$$

$$f_2(x) = x \tag{18}$$

where $f_1(x)$ and $f_2(x)$ are selected to be the "log-sigmoidal" and "pure-linear" transfer functions attached to the nodes in hidden layer and output layer respectively. The parameter $\varphi$, controls the steepness between two asymptotic values between 0 and 1.

Figure 4 shows the comparison between the targeted and predicted $k$-values from the ANN model. A statistical analysis is performed to determine the coefficient of determination ($R^2$) associated with the predicted $k$-values.

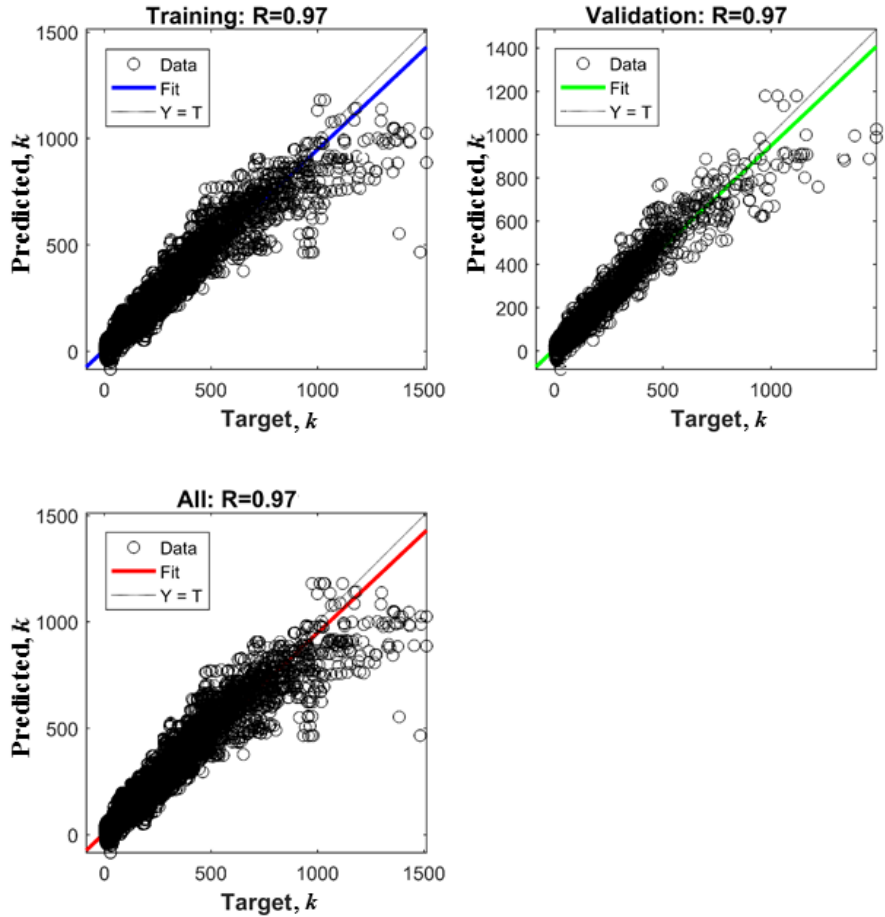

**Figure 4. Comparison of target versus predicted *k*-values using ANN model for training, validation, and overall datasets**

The high $R^2$ values of 0.97 for training and validation datasets indicate a good prediction accuracy of ANN model. Therefore, the developed ANN model is used to predict modified *k*-value for performance analysis of rigid pavements at any given layer structure, modulus and degree of bonding.

## 3. Validation of ANN predicted *k*-values

To examine the prediction accuracy of the developed ANN model, the *k*-values obtained as model output are used in the FE model of pavement structure with a Winkler foundation model and therefore the simulated deflection patterns are compared with the deflections of the full pavement model. The development of the FE model of full pavement structure is presented in the previous section. Figure 1a illustrates the schematic model of a full pavement structure and 1b shows the axisymmetric model of full pavement in ABAQUS. The simplified Winkler model is developed by characterizing the subgrade resistance with a series of springs and coupled with an equivalent PCC slab layer. The pavement section data are collected as inputs for the FE model, which are elaborated in the next subsection.

### 3.1. Data collection

The pavement sections data used in this study are collected from the LTPP database. In addition to the pavement layer structure, the database includes the modulus value for each pavement layer. The slab-base interface bonding ratios are calculated using the approach described in the reference [8]. Table 5 lists the slab and base thicknesses, slab, base and subgrade moduli and slab-base interface bonding ratio of each pavement section. To consider the climate impact, two pavement sections are selected from each of the four principal climatic zones.

**Table 5. Selected LTPP pavement sections**

| Climate zone | State | State Code | SHRP ID | Slab thickness (in) | Base thickness (in) | Backcalculated values | | | $\delta$ |
| --- | --- | --- | --- | --- | --- | --- | --- | --- | --- |
| | | | | | | Slab modulus (psi) | Base modulus (psi) | Subgrade modulus (psi) | |
| Wet-Freeze | Minnesota | 27 | 4034 | 10 | 3.6 | 6950000 | 22000 | 28000 | 0.52 |
| | Kentucky | 21 | 4025 | 9.8 | 6 | 6192000 | 92000 | 33000 | 0.9 |
| Wet-Nonfreeze | Alabama | 01 | 0606 | 10.3 | 6.3 | 7890000 | 22000 | 47000 | 0.5 |
| | North Carolina | 37 | 5037 | 7.8 | 15.1 | 5054000 | 326000 | 16000 | 0.11 |
| Dry-Freeze | Colorado | 08 | 7776 | 10.7 | 15.3 | 3748000 | 44100 | 26700 | 0.48 |
| | North Dakota | 38 | 3006 | 8.5 | 3.8 | 6100000 | 30000 | 24000 | 0.37 |
| Dry-Nonfreeze | New Mexico | 35 | 3010 | 7.9 | 6.9 | 7171000 | 68000 | 31000 | 0.22 |
| | Arizona | 04 | 0214 | 8.3 | 6.1 | 7994000 | 70000 | 23700 | 0.32 |

### 3.2. Comparison of deflection patterns

In this study, the deflection basin obtained from a pavement structure consisting of slab and base layers on a subgrade *k*-value (Winkler model) are compared with the deflections using the full pavement structure. Figure 5a shows a typical rigid pavement structure which consists of a PCC surface layer, and an unbound aggregate base course along with a subgrade foundation with the uniform *k*-value. Similar to the full pavement structure model, the Winkler model pavement structure is also subjected to an FWD load of 40 kN. The deflections from FEM analysis are obtained at 0, 0.3 m, 0.6 m and 0.9 m away from the loading point. Figure 4b presents the axisymmetric finite element mesh of an equivalent slab layer with the Winkler subgrade model. The thickness of the equivalent slab is calculated in Equations 19 and 20 [8].

$$I_{tr} = I_{slab} + I_{base} + \delta \sum A_i \bar{d}_i^2 \tag{19}$$

$$h_{eq} = \sqrt[3]{\frac{I_{tr}*12}{b}(1-\upsilon^2)} \tag{20}$$

where $I_{slab}$, $I_{base}$, and $I_{tr}$ are the moment of inertia of the slab, base and transformed pavement sections respectively; $A_i$ is the area of the slab and the transformed area of the base course; $\bar{d}_i$ is centroidal distance to each of the areas; $\delta$ is the interface bonding ratio; $h_{eq}$ is the equivalent slab thickness; and ν is the Poisson's ratio.

    The accuracy of the pavement response using the Winkler model primarily depends on the proper characterization of subgrade reaction (*k*-value) in the FE model. Many existing studies have investigated the problem of the slab-on-subgrade *k*-value using the FE method. A soil-structure interaction has been modeled in ABAQUS with spring stiffness representing the subgrade reaction *k*-value [29]. Recently, a fully coupled 3D train-track-soil model have been developed in a FE program and examined the capability of the model at high train speed [30-31]. The spring elements available in ABAQUS are utilized in that model to represent the ballast between the ties and the ground. Similarly, a Winkler foundation is approximated as a spring element by to simulate the dynamic vehicle interaction of a slab-track system [32].

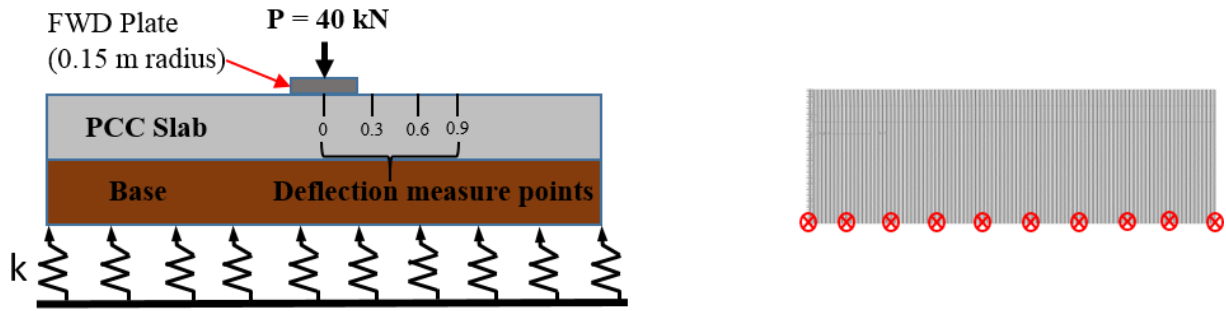

**Figure 5. (a) Schematic plot of a multi-layered pavement structure resting on Winkler foundation; (b) Axisymmetric model of equivalent slab with uniform spring stiffness in ABAQUS**

In this study, the SPRING1 elements in ABAQUS are used to model the actual physical spring with a *k*-value. The SPRING1 element, as shown in Figure 5b, acts between a node in the slab and ground. The linear spring behavior is defined by the modified *k*-value obtained from the ANN model as spring stiffness.

Figure 6 shows the comparison between deflection pattern calculated using FE analysis of full pavement structure model and Winkler pavement structure model. Overall, the deflection patterns obtained from the simplified Winkler model are in good agreement with the full structure model for all the selected pavement sections. A small deviation is observed in the furthest region away from the loading point. This is mainly caused by the boundary effect on the calculated deflections for the simplified Winkler model. The high accuracy of deflection patterns also indicates that the predicted *k*-value from the ANN model can take account of the slab-base interface condition at any given degree of bonding.

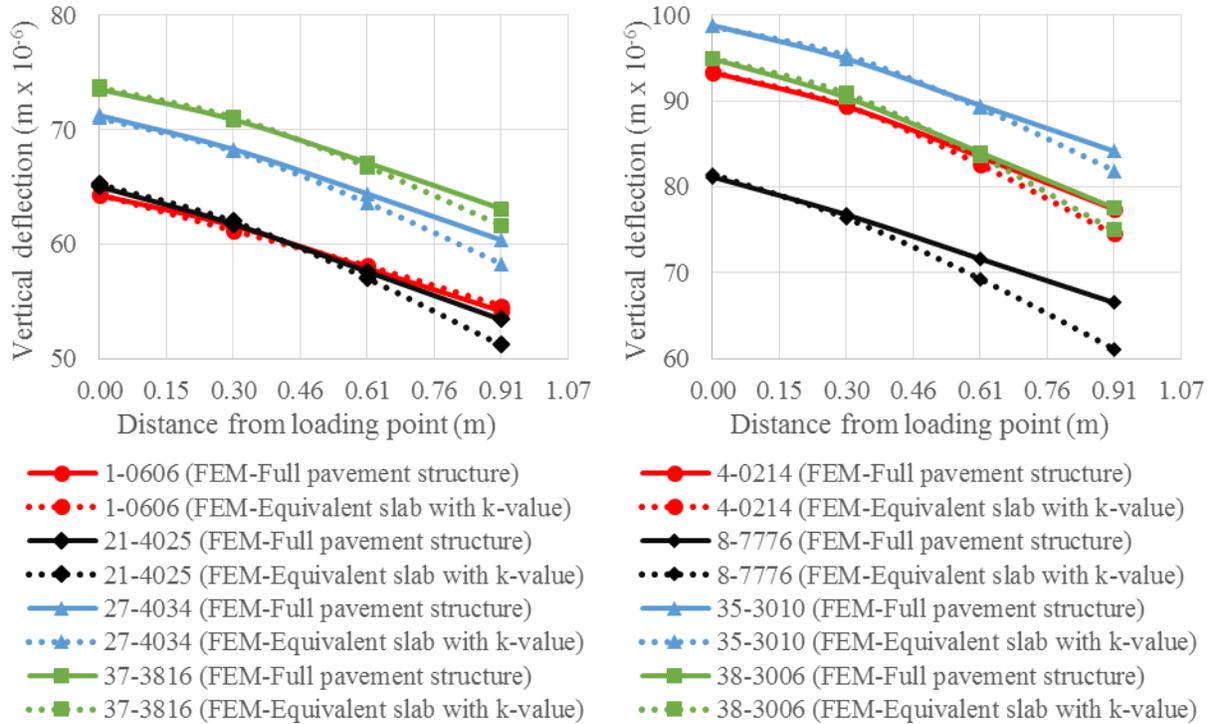

**Figure 6. Comparison of vertical deflections due to FWD loading obtained from full pavement structure model and equivalent slab-modified k-value model simulations**

## 4. Sensitivity of water content and degree of bonding on *k*-value

To investigate the effect of base water content on subgrade *k*-value and pavement performance, $M_R$ values of base material are determined at the three different water contents. Table 6 lists the $M_R$ values of base layer for each pavement structure at the three selected water contents: (1) saturated volumetric water content, (2) equilibrium volumetric water content, (3) 80% of equilibrium volumetric water content. Table 6 also list three different bonding conditions for each pavement structure: (1) no bond, (2) partially bonded, and (3) fully bonded. Each bonding condition is an input in the developed ANN model and compared with the predicted *k*-values from Pavement ME design.

### 4.1. Estimation of equilibrium volumetric water content at base layer

As it has been discussed earlier, the mean volumetric water content at base and subgrade layer is determined by the equilibrium suction at a certain depth of moisture active zone. To determine the equilibrium suction value at the depth of moisture active zone for the selected pavement locations, this study used the equilibrium suction map developed by Saha et al. [19]. Figure 7 illustrates the schematic profile of equilibrium suction profile underneath pavement surface with depth. The slope of suction vs. depth profile in base/subgrade layer is derived from the formulation of hydraulic head. According to the Bernoulli's Principle, the total hydraulic head is composed of pressure head and elevation head [33].

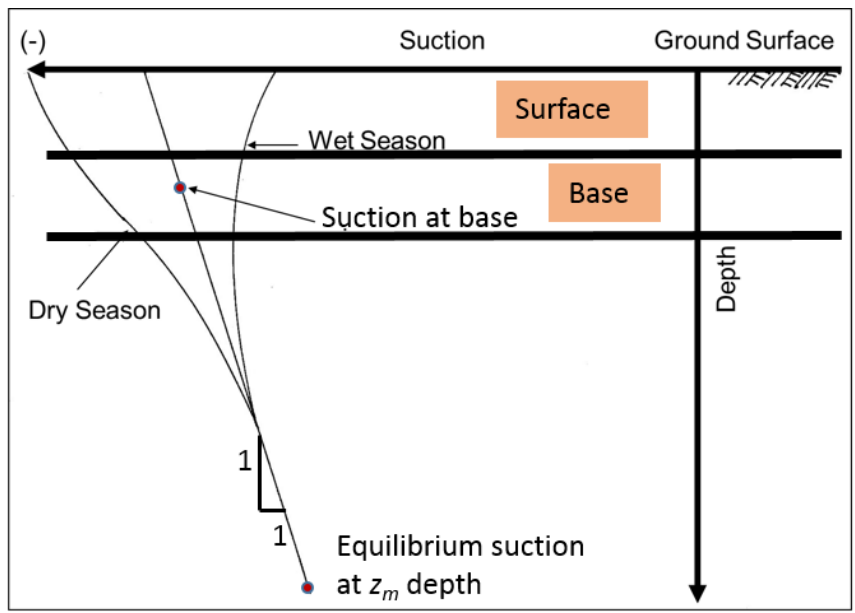

**Figure 7. Schematic illustration of in-situ suction at mid depth of base layer**

$$H = h + Z \tag{21}$$

where *H* is the hydraulic head (unit: cm), *h* is the suction pressure head, in terms of the elevation difference of the water column (unit: gm-cm/cm); and, Z is the elevation (unit: cm)

The hydraulic gradient is expressed in Equation 21,

$$\frac{\partial H}{\partial Z} = \frac{\partial h}{\partial Z} + 1 \tag{22}$$

However, the flow discharge, *Q* can be expressed as a function of permeability (*k*) and hydraulic gradient ($\frac{\partial H}{\partial Z}$) following the Darcy's law, as shown in Equation 23.

$$Q = -kA \frac{\partial H}{\partial Z} \tag{23}$$

Again, $Q = A * v$ \hfill (24)

where, A is the area and, v is the velocity

From Equations 23 and 24, it is deduced that

$$v = -k \frac{\partial H}{\partial Z} \tag{25}$$

In equilibrium condition the bulk motion of water flow is zero,

$$v = 0 \Rightarrow \frac{\partial H}{\partial Z} = 0 \tag{26}$$

Substituting Equation 26 into Equation 22 yields,

$$\frac{\partial h}{\partial Z} = -1 \qquad (27)$$

Therefore, the suction profile maintains a 1 by 1 slope when both suction and depth are expressed in the unit of cm. Using the elevation difference between the depth of moisture active zone and base mid and equilibrium suction value at moisture active zone depth, mean suction value are calculated at the base layer. Subsequently, the mean volumetric water content at base layer are calculated using the SWCC equation developed by Fredlund and Xing [34].

$$S = C(h) \times \left[ \frac{1}{\{\ln[e + (\frac{h}{a_f})^{b_f}]\}^{c_f}} \right] \qquad (28)$$

where $h$ represents the matric suction in the unit of cm of water pressure and $C(h)$ is a correction factor defined as

$$C(h) = 1 - \frac{\ln(1 + \frac{h}{h_r})}{\ln[1 + (\frac{1.021 \times 10^7}{h_r})]} \qquad (29)$$

The four fitting parameters i.e., $a_f$, $b_f$, $c_f$ and $h_r$ in Equations 28 and 29 are predicted using the ANN model developed by Saha et al. [24]. Soil physical properties such as gradation, Atterberg limits, saturated volumetric water content ($\theta_{sat}$) and mean annual air temperature are used as input parameters, and the fitting parameters are obtained as output. The coefficients $k_1$, $k_2$ and $k_3$ in $M_R$ model are also predicted using another ANN model developed by Saha et al. [9]. Table 6 lists the predicted SWCC fitting parameters and $M_R$ model coefficients for the selected pavement sections. Using the predicted coefficients, $M_R$ values are calculated for the base materials at three different water contents.

**Table 6. Selected range of moisture and degree of bonding for each LTPP sections**

| State Code | SHRP ID | $\delta$ | Saha et al. (2018a) | | | | $\theta$ | S (%) | -hm (kPa) | f | Saha et al. (2018b) | | | $M_R$ (kPa) |
|---|---|---|---|---|---|---|---|---|---|---|---|---|---|---|
| | | | $a_f$ | $b_f$ | $c_f$ | $h_r$ | | | | | $k_1$ | $k_2$ | $k_3$ | |
| 27 | 4034 | 0 | 0.52 | 4.91 | 2.62 | 1.65 | 3000 | 0.009 | 5.14 | 315 | 1 | 689.3 | 0.66 | -0.03 | 24519 |
| | | | | | | | | 0.011 | 6.28 | 237 | 1 | | | | 23890 |
| | | 1 | | | | | | 0.174 | 100.00 | 1.43 | 5.78 | | | | 20503 |
| 21 | 4025 | 0 | 0.9 | 5.86 | 0.34 | 1.74 | 2999 | 0.065 | 39.85 | 751 | 1 | 945.55 | 0.67 | -0.29 | 280163 |
| | | | | | | | | 0.081 | 49.66 | 201 | 1 | | | | 155712 |
| | | 1 | | | | | | 0.163 | 100.00 | 0.0006 | 7.35 | | | | 70092 |
| 01 | 0606 | 0 | 0.5 | 6.71 | 1.01 | 0.07 | 2998 | 0.120 | 76.20 | 14092 | 1 | 913.7 | 0.73 | -0.03 | 1808022 |
| | | | | | | | | 0.150 | 95.25 | 159 | 4.99 | | | | 264142 |
| | | 1 | | | | | | 0.157 | 100.00 | 0.014 | 6.35 | | | | 15359 |
| 37 | 5037 | 0 | 0.11 | 7.57 | 0.98 | 1.08 | 2999 | 0.07 | 37.68 | 473 | 1 | 431.43 | 0.92 | -0.23 | 99425 |
| | | | | | | | | 0.087 | 46.84 | 247 | 1 | | | | 70929 |
| | | 1 | | | | | | 0.185 | 100.00 | 0.012 | 5.38 | | | | 14665 |
| 08 | 7776 | 0 | 0.48 | 1.06 | 1.01 | 0.69 | 2999 | 0.073 | 27.12 | 3329 | 1 | 983.52 | 0.207 | -0.027 | 162921 |
| | | | | | | | | 0.091 | 33.81 | 731 | 1 | | | | 126136 |
| | | 1 | | | | | | 0.269 | 100.00 | 0.002 | 3.71 | | | | 74777 |
| 38 | 3006 | 0 | 0.37 | 1.00 | 1.01 | 0.79 | 2999 | 0.041 | 23.68 | 2252 | 1 | 544.43 | 0.65 | -0.08 | 140457 |
| | | | | | | | | 0.052 | 30.04 | 593 | 1 | | | | 75007 |
| | | 1 | | | | | | 0.172 | 100.00 | 0.002 | 5.78 | | | | 25624 |
| 35 | 3010 | 0 | 0.22 | 5.30 | 3.35 | 1.05 | 2998 | 0.014 | 6.45 | 1717 | 1 | 859.5 | 0.73 | -0.025 | 85988 |
| | | | | | | | | 0.017 | 7.83 | 851 | 1 | | | | 64877 |
| | | 1 | | | | | | 0.217 | 100.00 | 2.827 | 4.6 | | | | 34969 |
| 04 | 0214 | 0 | 0.32 | 5.05 | 0.12 | 2.18 | 2999 | 0.052 | 33.08 | 12438 | 1 | 900.14 | 0.509 | 0.047 | 359581 |
| | | | | | | | | 0.065 | 41.35 | 2285 | 1 | | | | 172806 |
| | | 1 | | | | | | 0.157 | 100.00 | 0.0006 | 6.36 | | | | 33750 |

Figure 8 presents the sensitivity of degree of bonding on the subgrade *k*-value using the developed ANN model and the Pavement ME design model. Both models show similar range of sensitivities to the degree of bonding on the *k*-value. However, as shown in Figure 8b, the Pavement ME design model has no partial bonding condition whereas the ANN model can predict the *k*-value at any bonding condition. The percentage change in *k*-value at partial and full bond condition are computed by using Equation 30.

$$k-value \text{ change}_{(partial/full\ bond)}(\%) = \frac{k-value_{(partial/full\ bond)} - k-value_{(no\ bond)}}{k-value_{(no\ bond)}} * 100 \qquad (30)$$

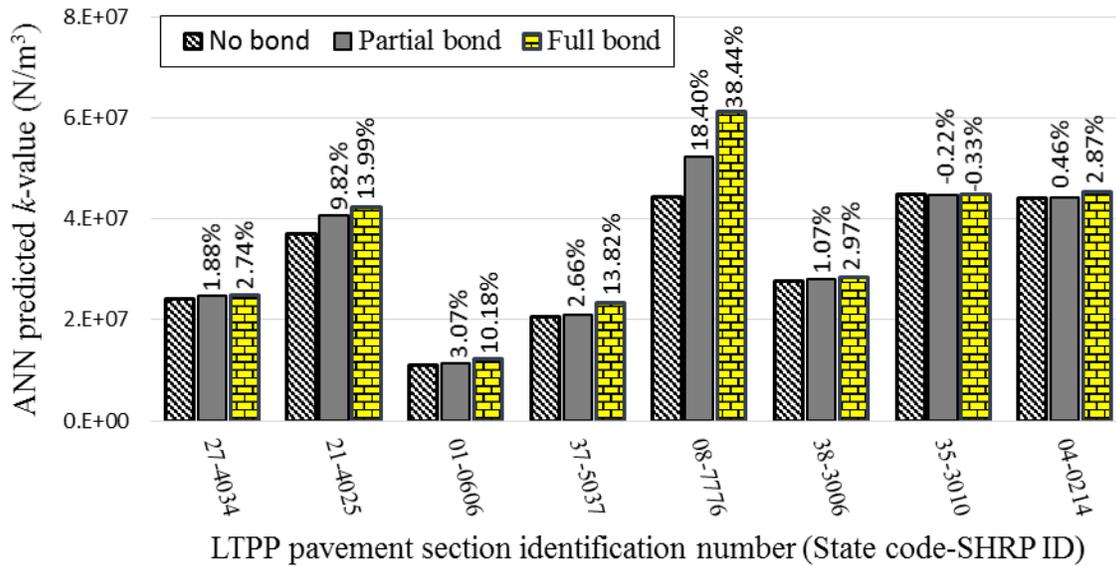

(a)

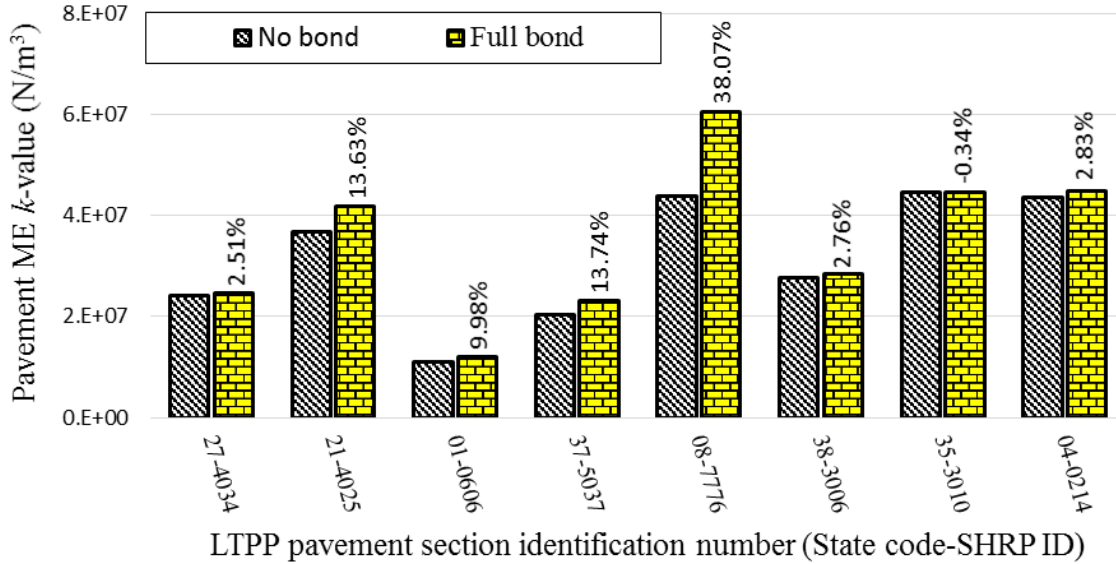

(b)

Figure 8. Sensitivity of slab-base interface bonding on *k*-value using (a) ANN model; (b) Pavement ME design model

LTPP sections 21-4025 and 08-7776 show a significant increase in the predicted *k*-value due to the increase in degree of bonding. This is mainly because they have a higher modulus ratio ($E_{base}/E_{slab}$) compared to other sections which contribute to greater impact on the thickness of equivalent section and therefore predicted *k*-value.

Figure 9 compares the sensitivity of moisture on the *k*-value using the ANN model and Pavement ME design model. The *k*-value in Pavement ME design has extremely low sensitivity to the moisture of base material. As a contrast, the proposed ANN model shows relatively higher sensitivity for all the selected pavement sections. The percentage change in k-value at equilibrium water content and 80% of equilibrium water content are computed based on the k-value using saturated water content on base layer, as expressed below

$$k-value\ \text{change}_{(Eq.vol.wc./80\%\ Eq.vol.wc.)}(\%) = \frac{k-value_{(Eq.vol.wc./80\%\ Eq.vol.wc.)} - k-value_{(Sat.vol.wc.)}}{k-value_{(Sat.vol.wc.)}} *100 \quad (31)$$

The percentage change in *k*-value using ANN model are 2 to 5 times higher than the results from Pavement ME design model for identical change of water content in base layer. This is mainly because both water content and the corresponding matric suction is included in the calculation of the modified $M_R$ value which later used as an input in the ANN model. Therefore, greater change in $M_R$ value causes greater change in k-value for all the selected pavement sections.

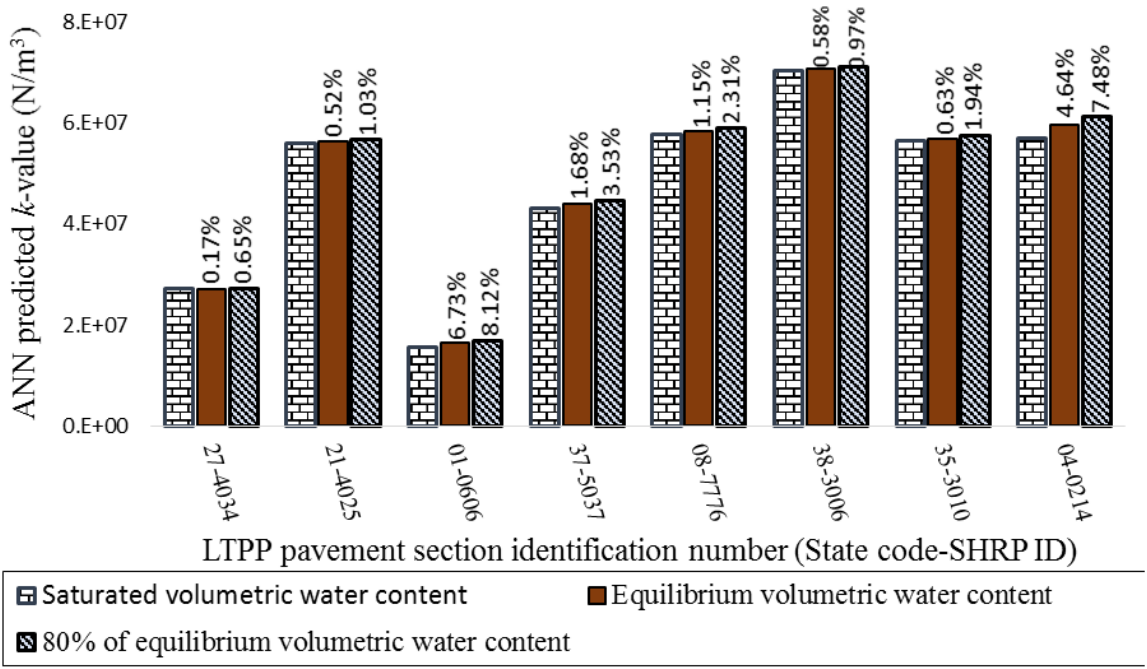

(a)

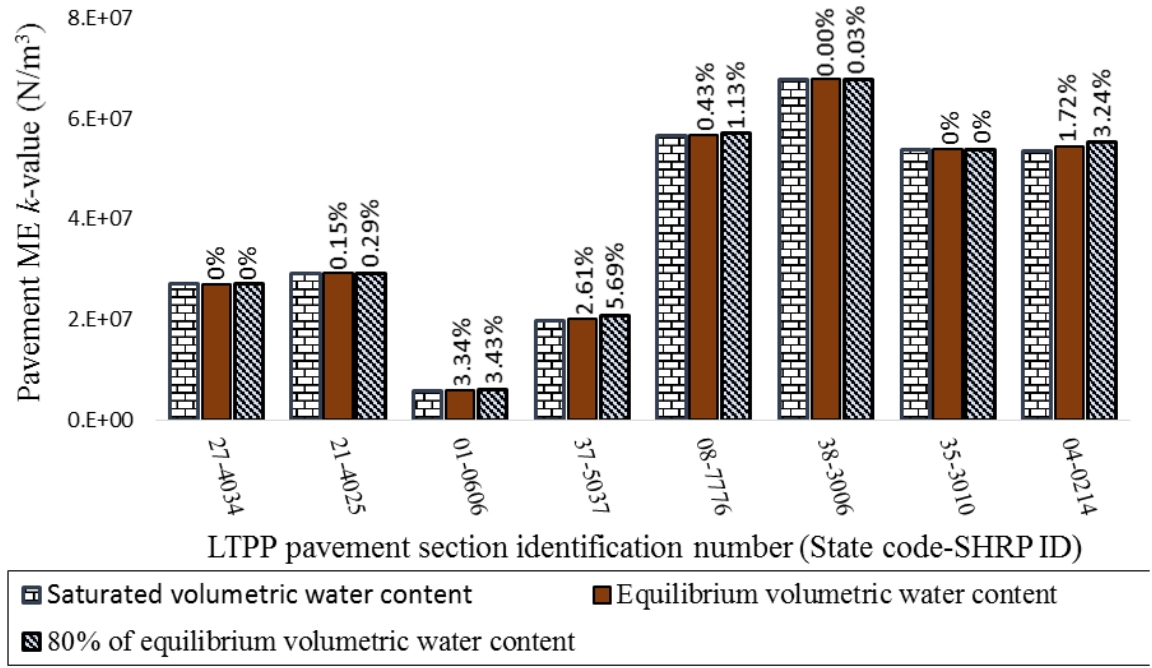

**(b)**

**Figure 9. Sensitivity of base material moisture on *k*-value using (a) ANN model; (b) Pavement ME design model**

Although the combination of moisture sensitive $M_R$ model and the developed ANN model shows high sensitivity to moisture and slab-base degree of bonding on *k*-value, it is still necessary to evaluate the effects of moisture and degree of bonding on pavement performance. The next section describes the prediction of the fatigue cracking and faulting performance for various bonding and moisture conditions and compares the results with those predicted by the Pavement ME design model.

## 5. Prediction of cracking and faulting distress performance

The Jointed Plain Concrete Pavement (JPCP) performance in Pavement ME design is evaluated in terms of three distress indicators, i.e., transverse cracking, joint faulting and roughness. Both transverse cracking and faulting performance are considered in this study to evaluate the effect of water content in base layer and slab-base interface bond ratio.

Transverse cracking in rigid pavement initiates either from the pavement surface (top-down) or the slab bottom (bottom-up) depending on the combined effect of traffic and environmental loadings. A fatigue damage related transverse cracking model is used in the Pavement ME design for both bottom-up and top-down cracking, as shown in Equation 32

$$CRK_{Bottom-up/Top-down} = \frac{1}{1+FD^{-1.68}} \tag{32}$$

where *FD* is the fatigue damage caused by the traffic and environmental loadings. The total cracking amount in JPCP is calculated using Equation 33.

$$TCRACK = (CRK_{Bottom-up} + CRK_{Top-down} - CRK_{Bottom-up} * CRK_{Top-down}) * 100\% \tag{33}$$

The fatigue damage in the pavement is predicted using Miner's damage accumulation law.

$$FD = \sum \frac{n_{i,j,k,l,m,n}}{N_{i,j,k,l,m,n}} \tag{34}$$

Herein,

$$\log(N_{i,j,k,l,m,n}) = C_1(\frac{MR_i}{\sigma_{i,j,k,l,m,n}})^{C_2} + 0.4371 \tag{35}$$

where $n_{i,j,k,l,m,n}$ and $N_{i,j,k,l,m,n}$ denote the applied and allowable number of load applications respectively over a design period, *and i, j, k, l, m,* and *n* account for age, month, axle type, load level, temperature difference and traffic path respectively. Here, the allowable number of load applications is calculated based on the ratio of PCC modulus of rupture and the applied stress at slab bottom or top.

Similarly, an incremental damage accumulation approach is adopted by the Pavement ME design to predict joint faulting, as expressed in Equations 36-37.

$$Fault_m = \sum_{i=1}^{m} \Delta Fault_i \tag{36}$$

$$\Delta Fault_i = C_{34} * (FAULTMAX_{i-1} - FAULT_{i-1})^2 * DE_i \tag{37}$$

where $Fault_m$ is the mean joint faulting at the end of month m; $\Delta Fault_i$ is the incremental change (monthly) in mean joint faulting during month *i*; $FAULTMAX_i$ is the maximum mean joint faulting for month *i*; $DE_i$ is the differential deformation energy accumulated during month *i*; The differential deformation energy, $DE_i$, in Equation 37 is calculated using the difference between square of the slab corner deflections, as expressed in Equation 38.

$$DE = k/2(\delta^2_{loaded} - \delta^2_{unloaded}) \tag{38}$$

where $\delta_{loaded}$ and $\delta_{unloaded}$ are the loaded and unloaded corner deflection respectively.

The Pavement ME design guide adopts an incremental distress calculation procedure which requires a great number of computations of critical stresses and deflection basin to estimate the monthly cracking and faulting damage (i.e., different loads, load positions and temperature gradients) over a design period. Both $\sigma_{i,j,k,l,m,n}$ and ($\delta_{loaded}$ - $\delta_{unloaded}$) in Equation 35 and 38 respectively are computed for different axle types of various load levels that passes through the traffic path under each climatic condition i.e., age, season, temperature difference. Due to the unavailability of traffic, climate and age condition data for each LTPP pavement sections, this study aims to develop a comparative study of critical pavement responses at a fixed loading, climatic and age condition which further project the sensitivity of pavement performances.

## 5.1. Sensitivity of water content and interface bonding on critical pavement response

The fatigue damage at the slab occurs due to a critical combination of heavy traffic and environmental loadings. When the pavement is exposed to a high negative temperature gradient, the PCC slab curls upward. In addition to that if axle loads are applied at opposite ends of the slab simultaneously, a high tensile stress generates at the top of the slab and initiate top-down cracking [11]. Similarly, bottom-up transverse cracking initiates when the tensile bending stress is maximum at the bottom of the slab. The critical traffic axle position for bottom-up cracking is determined as near the longitudinal edge of the slab, halfway between transverse joints. The presence of a high temperature gradient through the slab intensifies the tensile stress at slab bottom heavily [35]. Faulting distress in JPCP is primarily defined as the difference in elevation across transverse joint. Therefore, the application of repeated heavy axle loads crossing transverse joints is considered as critical condition for joint faulting.

In this study, to compute the critical stress for top-down cracking, two single axles load, weighing 22,000 lbs. each, are applied at opposite ends of the slab on the transverse joints in addition to a negative $3^0C$ temperature gradient. The condition for bottom-up cracking is one single axle load of 22,000 lbs. midway between the transverse joints when slab was exposed to a positive $3^0C$ temperature gradient. Lastly, one tandem axle load, weighing 57,000 lbs. near the edge of the transverse joint is applied for the computation of joint faulting. No temperature gradient is applied for the faulting distress calculation. The ISLAB2000 finite element (FE) software is used to accurately compute the stress and deflection responses for rigid pavements.

### 5.1.1. Interface bond sensitivity

Figure 10 shows the effects of the degree of bonding on pavement responses i.e., tensile stress at surface, tensile stress at the slab bottom and differential deflection across the transverse joint using the *k*-value predicted from the ANN model and the Pavement ME design model.
It is observed that the developed ANN model has larger sensitivity of tensile stress and differential deflection due to change in degree of bonding at the slab-base interface. For both the ANN model and the Pavement ME design model, the fully bonded condition shows the lowest tensile stress and differential deflection whereas no bonding between slab and base yields the highest tensile stress and differential deflection. It is also seen that the Pavement ME design model can only calculate tensile stress and deflection at two extreme bonding conditions but the developed ANN model has the capability of predicting tensile stress and deflection at partially

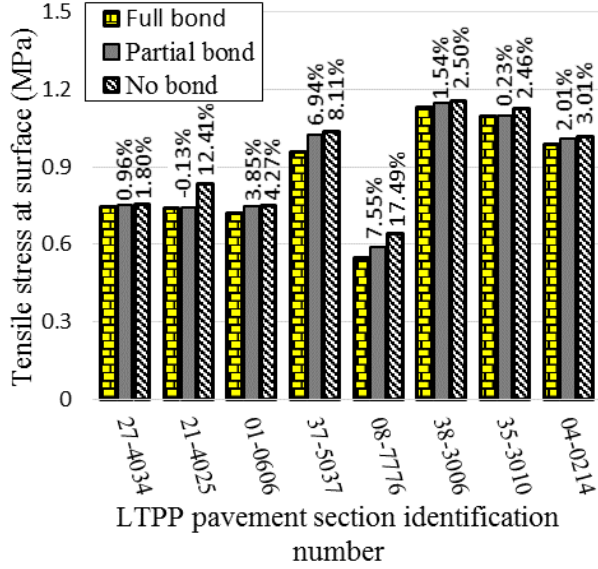 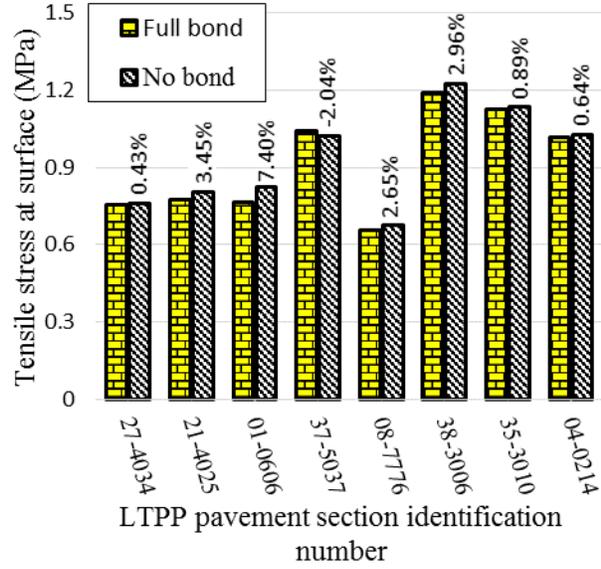

**ANN model**          **Pavement ME design**

**(a)**

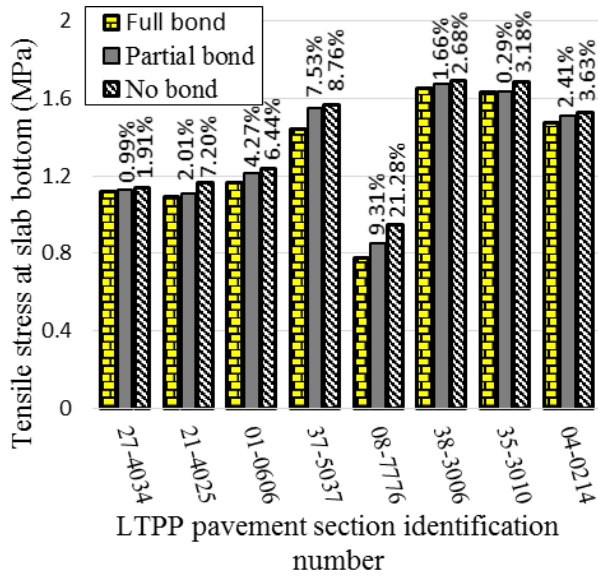 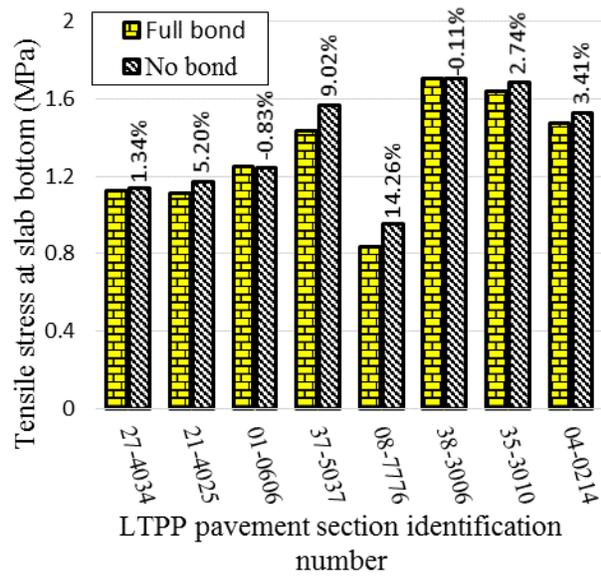

**ANN model**          **Pavement ME design**

**(b)**

bonded conditions as well. Therefore, more accurate prediction of stress and deflections at critical locations enable precise computations of pavement distresses i.e., cracking and faulting at partially bonded conditions which contribute to more economic design of rigid pavement.

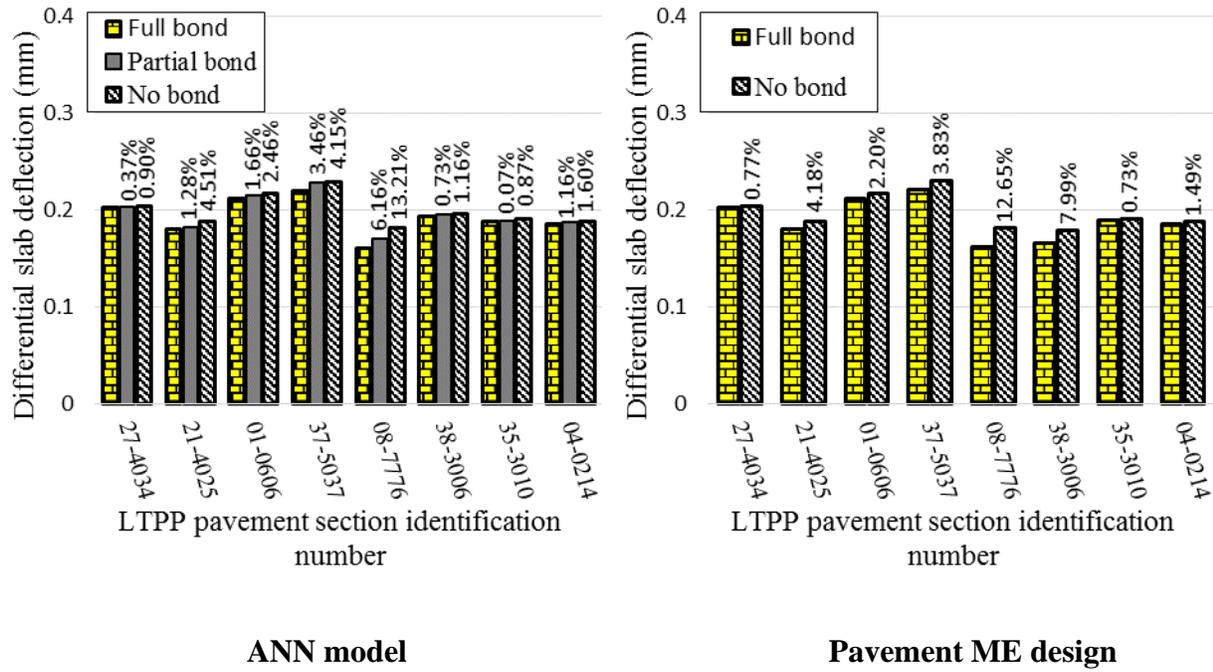

**(c)**

**Figure 10. PCC slab-base interface bond sensitivity on (a) tensile stress at top of slab; (b) tensile stress at bottom of slab; and (c) differential deflection on transverse joints**

### 5.1.2. Water content sensitivity

Figure 11 shows the sensitivity of water content on tensile stress and differential deflection using the proposed water content and suction dependent $M_R$ model in the base course and the subgrade together with the ANN model and the Pavement ME design model.

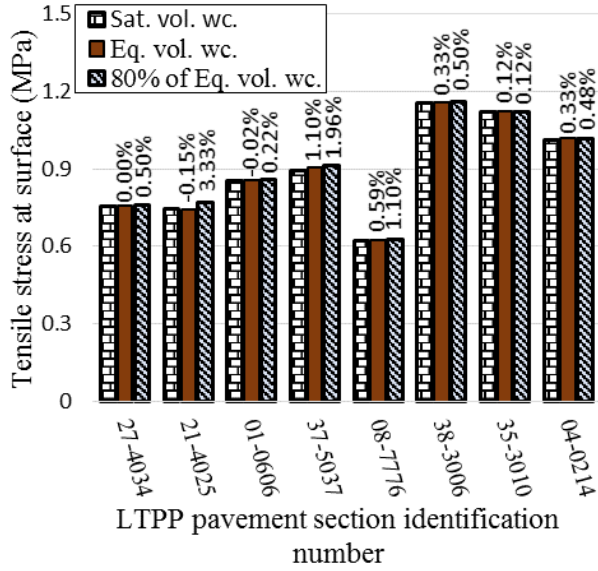
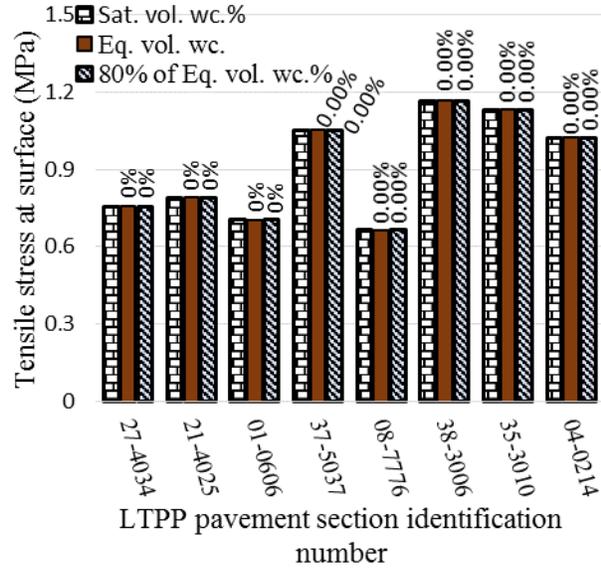

**ANN model**          **Pavement ME design**

(a)

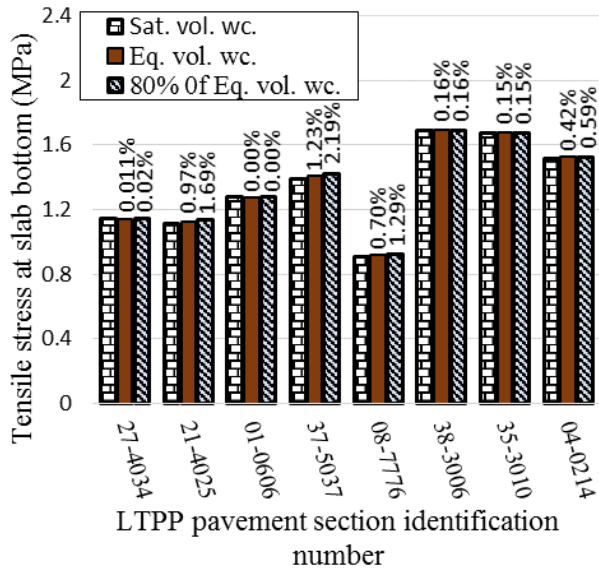
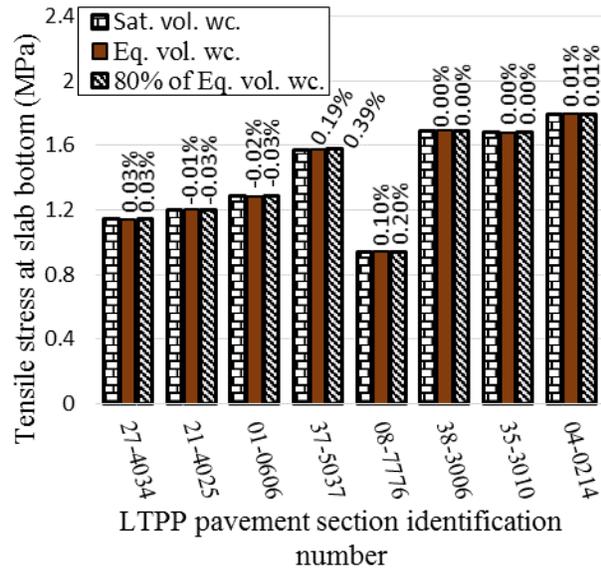

**ANN model**          **Pavement ME design**

(b)

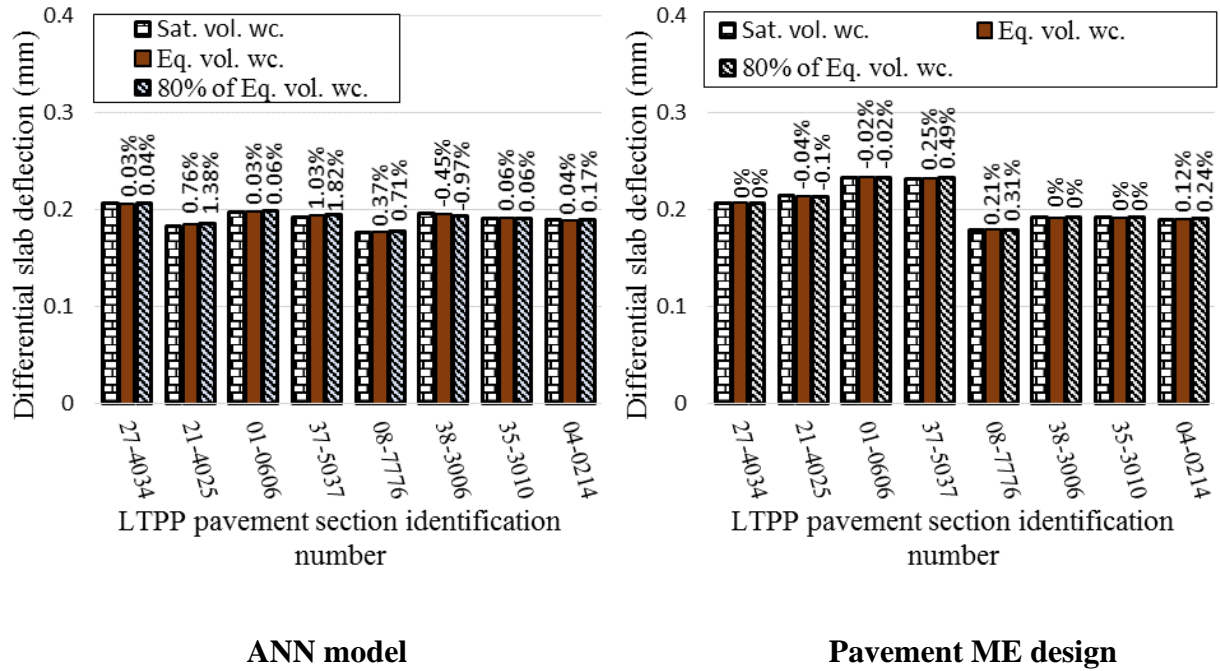

**ANN model**          **Pavement ME design**

(c)

**Figure 11. Base layer moisture sensitivity on (a) tensile stress at top of slab; (b) tensile stress at bottom of slab; and (c) differential deflection on transverse joints**

The ANN model has more reasonable sensitivity than the Pavement ME design model to the change of base water content in terms of the tensile stress at slab top and bottom and differential deflection. Tensile stress at the slab top and bottom show no sensitivity at all to the water content of base material whereas the combination of modified $M_R$ value and ANN model present significant change in the calculated tensile stresses for all the pavement sections. It is observed that the calculated tensile stresses on the pavement sections 08-7776 and 04-0214 have the greater sensitivity to the change of water content in base material. The difference of matric suctions in base materials of these pavement sections are much higher than other pavements due to the same range of saturation difference. Therefore the $M_R$ values changes significantly which contribute to the calculated stresses at critical locations.

Finally, the results of this study justify the fact that the modified models are capable of accurately predicting various pavement responses including stress and strain at the bottom and top of surface layer and differential deflection across the joint with different base properties under changing saturation conditions, while such differences cannot be predicted using the Pavement ME Design approach.

6. Summary and conclusions

This paper presents an ANN approach to predict the modified modulus of subgrade reaction ($k$-value) model by using the pavement layer thicknesses, moduli and interface bonding ratios. The ANN model has reasonable sensitivity to water content of base material and slab-base interface bonding ratio in terms of the predicted $k$-value. The major findings of this paper are summarized as follows.

- A three-layered neural network architecture consisting of one input layer, one hidden layer and one output layer were constructed for predicting the modified $k$-value. The input variables included slab and base thickness, slab, base, and subgrade modulus and slab-base interface bonding ratio. The hidden layer was assigned 20 neurons. A total of 27,000 simulation cases were modeled in the finite element program ABAQUS and the predicted deflection patterns were utilized to develop the ANN model.
- To consider the effects of water content in the base layer and the predicted $k$-value, the base $M_R$ value was calculated using a suction and water content dependent $M_R$ model. To determine an in-situ range of suction and water content, the equilibrium suction value was estimated first at a specific depth of the moisture active zone. Then the suction and

- the corresponding water content in the base layer were calculated using a 1 to 1 slope of the suction versus depth profile.
- The predicted *k*-values from the ANN model were compared against the k-values using the Pavement ME design approach. According to the ME design approach the *k*-value can only be estimated for a frictionless slip or fully bonded interface. As a contrast, the developed ANN model predicted *k*-values under any partially bonded conditions. The sensitivity of water content in the base layer was also examined on the predicted k-values using the developed ANN model and the Pavement ME Design model. Compared to the ME Design model, the developed ANN model had 3 to 5 times (by percentage) higher sensitivity of water content of the base layer for the selected pavement sections in this study.
- The effects of moisture of base material and slab-base interface bonding were also evaluated on the critical pavement responses i.e., tensile stress and differential deflection using a combination of moisture dependent $M_R$ model and the developed ANN model. The $M_R$ and *k*-values were input in the ISLAB2000 software and the computed results showed higher sensitivity than Pavement ME design on tensile stress at slab top due to moisture change in base layer. Similarly, the pavement ME design models showed almost zero sensitivity to moisture change in base layer on the computed tensile stress and deflections.

**Acknowledgements**

The authors acknowledge the financial support provided by the National Cooperative Highway. Research Program [Project Number: NCHRP 01-53].